\documentclass{article}

\usepackage{arxiv}

\usepackage[utf8]{inputenc} 
\usepackage[T1]{fontenc}    
\usepackage{hyperref}       
\usepackage{url}            
\usepackage{booktabs}       
\usepackage{amsfonts}       
\usepackage{nicefrac}       
\usepackage{microtype}      
\usepackage{lipsum}
\usepackage{graphicx}
\graphicspath{ {./images/} }

\usepackage{amsmath}
\usepackage{amssymb}
\usepackage{amstext}
\usepackage{amsthm}
\usepackage{chemarrow}
\usepackage{mathtools}
\usepackage{romannum}
\usepackage{tikz}
\usetikzlibrary{arrows,positioning,trees} 
\usepackage{subcaption}

\title{Physical System for Non Time Sequence Data}

\author{
 Xiongren Chen \\
  School of Computer Science\\
  University of Adelaide\\
  \texttt{xiongren.chen@adelaide.edu.au} \\
}

\begin{document}
\maketitle
\begin{abstract}
We propose a novelty approach to connect machine learning to causal structure learning by jacobian matrix of neural network w.r.t. input variables. In this paper, we extend the jacobian-based approach to physical system which is the method human explore and reason the world and it is the highest level of causality. By functions fitting with Neural ODE, we can read out causal structure from functions. This method also enforces a important acylicity constraint on continuous adjacency matrix of graph nodes and significantly reduce the computational complexity of search space of graph.
\end{abstract}


\section{Introduction}
As a financial quantitative for years, I always have to make predictions which generally have an assumed premise. For example, in the financial markets, if the U.S. dollar depreciates, how it affects the price of crude oil, whether it goes up or down, and then we make investments based on that prediction. It is a common practice to obtain historical price observations of the dollar and crude oil to calculate a correlation coefficient. The correlation coefficient and the change in the dollar are then used to calculate the change in the price of crude oil. From a statistical point of view, the correlation coefficient is a statistical indicator of how closely two variables are correlated and generally reflects the degree of linear correlation. A change in one variable can be obtained through the correlation coefficient for another variable. In the field of machine learning, the technique of learning the relationship between variables from data and then making predictions is very well established. However, we still need to be very cautious about using this technique in the financial field, as predictions based solely on correlations between data are not widely used stress tests in the financial market. For example, a typical stress test case would be if the central bank adjusts the interest rate, how does it affect a stock market index. This is where we have to make calculations using human expertise in the financial market, rather than simply using correlations. Human expertise in a particular field is generally presented in the form of differential equations, which in physical world can also be called physical systems. And the causal relationships between variables can be easily read out from inside the differential equations. In this work, we try to build differential equations by observational data given certain assumptions and constraints and then read out the causal relationships between variables from the physical systems \cite{Bernhard2019}.

Differential equations are widely used in various areas of modern science, such as the Black-Scholes option pricing model for the financial system, population development models and traffic flow models for the social sciences, and especially in physics, where they are used extensively in electromagnetic fluid dynamics, chemical fluid dynamics, power meteorology, ocean dynamics, and groundwater dynamics. As an example, R.M.Anderson gives an ordinary differential equation model of infectious disease dynamics\cite{Anderson1979},
\begin{equation}
\begin{split}
\frac{dX(t)}{dt} &:=  A - dX - \beta XY + \sigma Z, \\
\frac{dY(t)}{dt} &:=  \beta XY - (\gamma + \alpha + d)Y, \\ 
\frac{dZ(t)}{dt} &:=  \gamma Y - (\sigma + d)Z  \\
\end{split}
\end{equation}
Where $X(t), Y(t) and Z(t)$ denote the number of susceptible, infections and removed individuals respectively. And A denotes constant immigration rate, d is for constant natural death rate, $\beta$ represents transmission coefficient, $\alpha$ denotes disease-related death rate, $\gamma$ is for recovery rate and $\sigma$ represents loss of immunity rate. With differential equations, we can know the values of arbitrary variables in history, as well as predict future trends and changes in the system, or we can intervene with the system to get the desired results. At the same time, we can make inferences and give intuitive explanations, which is not possible with today's machine learning techniques. And, of course, we can easily read out causal relationships between variables. 

Usually, an Ordinary Differential Equation(ODE) has the form of,
\begin{equation} \label{eq:1}
\frac{d\boldsymbol{\mathrm{h}}(t)}{dt} :=  f(t,\boldsymbol{\mathrm{h}}(t)), \\
\end{equation}
with some known initial value, for example, $\boldsymbol{\mathrm{h}}(t = t_0) = \boldsymbol{\mathrm{h}}_0 $. If $f$ is Lipschitz, we can have a unique solution $\boldsymbol{\mathrm{h}}(t)$ according to The Picard–Lindelöf theorem\cite{Earl1955}. Equation \ref{eq:1} can also have the form as,
\begin{equation}  \label{eq:3}
\boldsymbol{\mathrm{h}}(t + dt) := \boldsymbol{\mathrm{h}}(t) + f(t,\boldsymbol{\mathrm{h}}(t)) dt, \\
\end{equation}
where $dt$ is the terms of infinitesimal differentials of time $t$. If we can get the solution to the ordinary differential equation, we can know which variables in the system affect the state at the next point of time. These variables can then have any direct causal influence on the result in the future. However, it is almost impossible to obtain ordinary differential equations and their solutions from large amounts of observational data and existing human expertise in a particular field, and random experiments and systematic interventions are generally required. 

In the era of machine learning, Neural ODE\cite{ode2018} takes inspiration from the following iterative process of ResNet\cite{resNet2016},
\begin{equation} 
\boldsymbol{\mathrm{h}}(t + 1) := \boldsymbol{\mathrm{h}}(t) + f(\boldsymbol{\mathrm{h}}(t)), \\
\end{equation}
This is equivalent to the Euler iterative solution of a differential equation\cite{Lu2017}. If we use more layers and smaller steps, it can be optimized to Equation \ref{eq:3}. That is the basic idea of Neural ODEs and function $f$ can be trainable neural networks. we can easily read out causal relationships between variables as a form of Jacobian matrix,
\begin{equation} \label{eu_eqn} J =
\begin{bmatrix} \frac{\partial f}{\partial x_1} \cdot \cdot \cdot \frac{\partial f}{\partial x_d}
\end{bmatrix} = 
\left[
\begin{array}{ccc}
  \frac{\partial f_1}{\partial x_1} & \cdots & \frac{\partial f_1}{\partial x_d} \\
   \vdots & \ddots & \vdots \\
   \frac{\partial f_d}{\partial x_1} & \cdots & \frac{\partial f_d}{\partial x_d}
\end{array}
\right]
\end{equation} 
\textbf{Contributions} The main contributions of this work can be summarized as follows,
\begin{itemize}
  \item We extend causal modeling to physical system which is usually in terms of ordinary differential equations. And physical systems can be seen as a full description of a dynamics system and ordinary differential equations can gain physical insight and explain functioning of a system.
  \item We use Jacobian matrix of function $f$ on input variables $x$ as causal relationships. Although not the first to propose this method, it is the first paper to use it in causal inference experiments.
  \item By comparing extensive experiments with current state-of-art methods for learning causal structures, the method in this paper wins in datasets with more dense causal relationships. It is shown that the method in this work is better suited to handle more complex causal relationships between nodes.
\end{itemize}
\section{From Statistical to Physical System}
\subsection{The Great Success of Statistical}
Probability theory relies on a probability space, the probability space ($\Omega,F,P$) totaling a measure of 1 (P($\Omega$)=1). The first term  $\Omega$ is a non-empty set, sometimes called the sample space. And the second term $F$ is a subset of the sample space $\Omega$ and ($\Omega,F$) together is called the probability measure space. The third term $P$ is called the probability, or probability measure\cite{Peter2017}. It is a function from the set $F$ to the real domain $R$. Each event is assigned a probability value between 0 and 1 by this function. For example, for the toss of a coin the sample space $\Omega$ is \{head, tail\}, $F$ is obtained from a random coin toss experiment, which may be $A = \{head\}$ or $B = \{tail\}$, and the corresponding probabilities P(A) = 0.5 and P(B) = 0.5. Probability theory allows us to infer the probability of the possible outcome of the next experiment from the data obtained from a historical random experiment. In general, we need to learn from historical random experiments to get the probability space, through which we know the possible distributions of the data, and the distribution obtained by learning can naturally give us the probability of different results of the next experiment. For example, an independent random experiment has a set of observations, $(x_1,y_1) \cdots (x_n,y_n)$, where $x_n$ is the input data and $y_n$ is the output data. We assume that $(x_n,y_n)$ are from variables $X$ and $Y$ which are independent and identically distributed(i.i.d.) with the unknown joint distribution $P_{XY}$. Generally existing machine learning and statistical methods follow the assumption that the data is i.i.d.. In machine learning, supervised learning is that we need to know $Y$ given a value of $X$ as the function $Y=f(x)$, or the probability of $Y$ given $X$ as $P(Y|X)$. Learning the decision function $Y=f(x)$ or the conditional probability distribution P(Y|X) directly from the data is typically used as a model for prediction, which we consider to be discriminative models. Typical discriminative models include K-Nearest Neighbors(KNN\cite{knn2009}), MultiLayer Perceptron(MLP), Decision Tree, Logistic Regression, Maximum Entropy Models, Support Vector Machine(SVM), Conditional Random Fields(CRFs\cite{crf2001}), etc. Another method is learning a join distribution $P(XY)$ through observational data, and then finding $P(Y|X)$ with $P(Y|X) = P(XY)/P(X)$, this method we called generative models. Typical generative models include the Hidden Markov Models(HMM), Mixed Gaussian models(MGMs), Averaged One-Dependence Estimators (AODE\cite{AODE2011}), Latent Dirichlet Allocation(LDA\cite{LDA2003}), and the Restricted Boltzmann Machine\cite{RBM2003}. 

The great success of deep neural networks in statistical methods is generally considered to be due to\cite{Bernhard2019}: (1) large amounts of data, especially precisely labeled data; (2) very powerful computational power, especially with the development of GPUs; (3) very complex and large computational systems with a large number of trainable parameters and (4) a closed static environment in which all data is assumed to be independent and identically distributed and the data distribution is constant. And since the existing deep learning models rely on i.i.d. data obtained in a closed environment, the models are working for some tasks but not for others. For example, if we add some noise to an image, the model may not be able to accurately identify and classify it. The same is true in the field of reinforcement learning, where a model trained in one game is difficult to transfer to another game because the model relies on a closed training environment and the i.i.d. data generated from the environment. If the environment changes or if human intervention occurs, the model will fail. For example, a set of i.i.d. data A=\{rain, not rain\} and B=\{floor is wet, floor is not wet\}. The model can learn from this set of data and go on to predict B from A, or predict A from B. However, if human intervention occurs and someone pours water on the floor causing the floor to be wet, then the previous model must fail. Open environment and systems interventions are not the realm of statistical but they are the realm of causal inference.

\subsection{Causal Graphical Models}
Reichenbach's common cause principle gives a clear explanation of the connection between statistical and causality\cite{Peter2017}: if two random variables $X$ and $Y$ are statistically dependent, then there exists a third variable $Z$ that affects both $X$ and $Y$. In other word, $Z$ screens $X$ and $Y$ from each other in the sense of that $X$ and $Y$ are independent of each other given $Z$. In the form of graph, there are three nodes $X$, $Y$ and $Z$ and two arrows with pointing from $Z$ to $X$ and from $Z$ to $Y$. $Z$ may coincide with either $X$ or $Y$, then there are only two points $X$ and $Y$ and one arrow in the graph. If $Z$ and $X$ coincide, then the arrow points from $X$ to $Y$. If $Z$ and $Y$ coincide, then the arrow points from $Y$ to $X$. For example, we have two random variables A=\{rain, no rain\} and B=\{floor wet, floor not wet\}, the corresponding causality is that A causes B and B cannot cause A. If we show causal relationships in the form of a graph, the nodes are A and B and the direction of the arrow is from A to B. But regardless of the causality, the observations are the same. Therefore, if we can't provide more information, we can't distinguish whether A affects B or B causes A. The information isn't more observational data but it's generally a stronger assumption on the data. More over, causality contains conditional independence properties which make the causal inference easier. For example, in the example of $Z$ causing both $X$ and $Y$, $X$ and $Y$ are independent of each other given $Z$. All these require a new kind of formalism to represent them. 

We use Directed Acyclic Graph with arrows pointing from parent(direct cause) node to child(direct effect) node as a formalism to represent causal relationships. These models are causal graphical models or graphical causal models which contains the observational data distribution and graph structure with nodes and arrows. We give its definition as follows,

\textbf{Definition of Causal Graphical Model} A Causal Graphical Model contains a Directed Acyclic Graph $\mathbb{G}(V,E) $ where $V$ is for nodes or vertices representing variables $X=(X_1,X_2,\cdots,X_n)$ and $E$ is for edges between nodes and a set of probability density function $P(X_j | X_{PA_j^\mathcal{G}})$ 
, such that the joint distribution $P(X)$ over $X$ equals the recursive product decomposition as follows\cite{Peter2017},
\begin{equation}  P(X) = \prod_d P(X_j | X_{PA_j^\mathcal{G}}) \end{equation} 
Where $X_{PA_j^\mathcal{G}}$ is for the parent nodes in DAG. This equation implies that variables $X_i$ is conditionally independent given the parent nodes of $X_i$. Causal Graphical Models can use do-calculus to intervene the system and have a new distribution  but they cannot answer counterfactual questions. Since this paper does not deal with interventions and counterfactuals, we skipped this part and if interested you can check out Peter's paper. The problem with Causal Graphical Models is that it is hard to make stronger restrictions on Causal Graphical Models to ensure identifiability. For example, decomposing $P(AB)$ can get $P(AB) = P(A)P(B|A)$ or $P(AB) = P(A)P(B|A)$ and we can't make other assumptions here to get the correct DAG. Therefore, we need to introduce Structural Causal Models(SCMs) or Structural Equation Models(SEMs), which can guarantee the identifiability after adding some restrictions on the functions.

\subsection{Structural Causal Models(SCMs) or Structural Equation Models(SEMs)}
\label{scms}

We give the definition of SEM as below.

\textbf{Definition of Structural Equation Model}  In a structural equation model over variables $X_1,X_2,\cdots,X_d$, there is a collection of $d$ equations(assignments):
\begin{equation}  X_j := f_j(X_{pa_j}, N_j) \hspace{1cm} j=1,...d, \end{equation} 
Where $X_{pa_j}$ is for the set of parent nodes of $X_j$ and $N_j$ is for mutually independent noise usually are Gaussian noise with zero mean. For example, we can get a SEM of rain and damp floors case we mentioned above,
\begin{equation} 
\begin{split}
A &:= N_1 \\
B &:= f_2(A) + N_2 
\end{split}
\end{equation}
SEM is based on data generative assumptions, which allows the addition of rich assumptions about how the data are generated, and thus the causal structure of the data can be obtained on the basis of function assumptions. SEMs are also the model basis for most current causal discovery methods. In a Causal Graphical Model, the decomposition of jointly distributed probabilities is difficult to distinguish between directions, such as $P(AB) = P(A) P(B|A) $or $P(AB) = P(B) P(A|B)$. It is also difficult to make assumptions over probabilities to ensure causal direction, since conditional probability and some simple continuous probability distributions are invertible. It is also difficult to distinguish directions in structural learning if noisy variables are not introduced in SEMs. For example, two random variables $X$ and $Y$ with relationship as $Y = 2X+1$ can be algebraically transformed to $X = (Y-1)/2$ . This symmetry is unintuitive in a causal relationship, since we cannot assume that it must be raining if the floor is wet, and we cannot assume that the air temperature has also changed by artificially adjusting the thermometer readings. In a SCM, we can also think of $X_{pa_j}$ as an endogenous variable, the noise variable $N_j$ as an exogenous variable for  unconsidered environmental factors, and there is only one exogenous variable. Endogenous variables are dependent on other variables and there is at least one edge pointing to the node; exogenous variables are independent of other variables and there is no edges pointing to the node. At the same time, assignment function $f_j$ can be linear or nonlinear. In the era of deep learning, it is easy to fit complex nonlinear functions with neural networks. Therefore, as a broadly used modeling framework, SCMs can generate a wide variety of powerful models to simulate complex data.

However, given a distribution $P_X$ on $X(X_1,X_2,\cdots,X_d)$, we can get different SEMs to entail this distribution. In the previous example of two variables, $P(AB)=P(A)P(B|A)$ can get a SEM or $P(AB)=P(A)P(B|A)$ can get another SEM but both point to $P(AB)$ at the same time. Therefore, we need additional information to help us get the right SEM, and this additional information would be the assumption of the data generation method $f_j$. We outline below several assumptions of $f_j$ to ensure identifiability results.

\subsubsection{Linear Non-Gaussian Acyclic Models}
Linear Non-Gaussian Acyclic Models(LiNGAM\cite{shimizu2006}) requires that the function $f_j$ in the assignment satisfy three conditions to ensure identifiability. First condition is that graph is a directed acyclic graph, in which the variable $X_1,X_2,\cdots,X_d$ has a sequential causal order and the preceding variables do not affect the following variables. Secondly, the model is linear which requiring the variables to be linear summations of the parent node variables in graph. The last condition is that the noise variables are non-Gaussian or there is only noise variable with Gaussian distribution. Further more, Noise variables are independent of other variables including noise variables. LiNGAM has the form of,
\begin{equation}  X_j := \sum_{k\in Pa_j} \beta_{jk}X_k + N_j\hspace{1cm} j=1,...d, \end{equation} 
Where all $N_j$ follow non-Gaussian distribution or only a $N_j$ is Gaussian distributed and all $\beta_{jk}$ are non-zero for all $k\in Pa_j$. Therefore, the SEM is identifiable from the joint distribution $P_X$. 

\subsubsection{Linear Gaussian Models with Equal Error Variances}
Linear Gaussian Models with Equal Error Variances(LGMEER\cite{LGMEER2014}) requires that the function $f_j$ in the assignment satisfy two conditions to ensure identifiability from the joint distribution over $X(X_1,X_2,\cdots,X_d)$: (1) the noise variables are Gaussian with variance $\sigma_2$ independent on $j$;(2) The model is linear which requiring the variables to be linear summations of the parent node variables in graph. LGMEER has the form of,
\begin{equation}  X_j := \sum_{k\in Pa_j} \beta_{jk}X_k + N_j\hspace{1cm} j=1,...d, \end{equation} 
Where all $\beta_{jk}$ are non-zero for all $k\in Pa_j$ and LGMEER is identifiable from the joint distribution $P_X$.

\subsubsection{Additive Noise Models(ANMs\cite{Peter2017})}
LiNGAM and LGMEER only solve the problem where the function is linear; in the nonlinear case, we generally assume an ANM which has the form of,
\begin{equation}  X_j := f_j(X_{pa_j}) +  N_j \hspace{1cm} j=1,...d, \end{equation} 
Where $X_{pa_j}$ is for the set of parent nodes of $X_j$ and $N_j$ is for mutually independent noise. An ANM with nonlinear assignments can ensure identifiable from the joint distribution $P_X$. If the assumption of Gaussian Noise $N_j$, then we have Nonlinear Gaussian Additive Noise Models which is also identifiable. If we have a stronger restriction on assignments $f_j$ with the form of,
\begin{equation} 
X_j := \sum_{k\in Pa_j} f_{jk}(X_k) + N_j\hspace{1cm} j=1,...d, 
\end{equation} 
Where all $f_{jk}$ are three times identifiable and nonlinear, then the model is a Causal Additive Model(CAM\cite{cam2014}).

\subsection{Physical systems or Ordinary Differential Equations}
SEMs can also be viewed in the form of differential equations. Let us first consider the case of discrete time in linear mode. There is an SEM over variables $X(X_1,X_2,\cdots,X_d)$ having following form,
$$
 X := WX +  N 
$$
Where $W$ is $d \times d$ adjacency matrix and $N$ represents noise vector. If $X$ is a sequence of variables $X^t$ having a value at time $t$, then we have iteration assignment,
$$
 X^t := WX^{t-1} +  N^{(t-1)} 
$$
As the linearity of the assignment, we have the form of the case of continuous time as,
$$
\frac{dX(t)}{dt} := C \\
$$
Where $C$ is constant matrix and we can certainly read out the causal relationships from $C$. For a nonlinear case, a SEM can be replaced by differential equations as,
$$
\frac{dX(t)}{dt} := f(X)
$$
or 
$$
X(t+\Delta t) := X_t + \Delta t \cdot f(X) 
$$
If we can get the solution to the ordinary differential equation, we can know which variables in the system affect the state at the next point of time. These variables can then have any direct causal influence on the result in the future and the causal relationships can be read out by Jacobian matrix of $f(X)$ on variable $X$. The various levels of causal modeling are summarized in Table \ref{tab:tab1} from Peter's paper. It is clear that the physical system is at the highest level and contains the most information. This paper aims to do causal modeling at the highest level and determine the validity of modeling by reading out the causal structure via Jacobian Matrix.

\begin{table}[]
\centering
\begin{tabular}{|l|c|c|c|c|c|}
\hline
\multicolumn{1}{|c|}{Model} & \begin{tabular}[c]{@{}c@{}}Predict \\ in i.i.d. \\ setting\end{tabular} & \begin{tabular}[c]{@{}c@{}}Predict under \\ changing distr. \\ or intervention\end{tabular} & \begin{tabular}[c]{@{}c@{}}Answer \\ counterfactual \\ questions\end{tabular} & \begin{tabular}[c]{@{}c@{}}Obtain \\ physical \\ insight\end{tabular} & \begin{tabular}[c]{@{}c@{}}Learn \\ from \\ data\end{tabular} \\ \hline
Physical System             & yes                                                                     & yes                                                                                         & yes                                                                           & yes                                                                   & ?                                                             \\ \hline
Structural causal model     & yes                                                                     & yes                                                                                         & yes                                                                           & ?                                                                     & ?                                                             \\ \hline
Causal graphical model      & yes                                                                     & yes                                                                                         & no                                                                            & ?                                                                     & ?                                                             \\ \hline
Statistical                 & yes                                                                     & no                                                                                          & no                                                                            & no                                                                    & yes                                                           \\ \hline
\end{tabular}
\caption{From paper \cite{Bernhard2019}: A summarization of different level of Causal Modeling. It is clear that the physical system is at the highest level and contains the most information.}
\label{tab:tab1}
\end{table}

\section{Neural ODEs for Causal Structure Learning}
\subsection{From ResNet to Neural ODEs}
Neural ODE\cite{ode2018} takes inspiration from the following iterative process of ResNet\cite{resNet2016},
$$
\boldsymbol{\mathrm{h}}(t + 1) := \boldsymbol{\mathrm{h}}(t) + f(\boldsymbol{\mathrm{h}}(t)), \\
$$
This is equivalent to the Euler iterative solution of a differential equation. If we use more layers and smaller steps, it can be optimized to Equation \ref{eq:3}. That is the basic idea of Neural ODEs and function $f$ can be trainable neural networks. We need to solve to equation and obtain the function $h(t)$ and its arguments $\theta$, so we use the conventional methods of solving ordinary differential equations, which starts solving the problem from the initial state $h_0$. This problem is generally called the initial value problem(IVP). Conventional methods for obtaining numerical solutions to differential equations by integrating the time variable include simple Euler methods and higher-order variants of the Runge-Kutta method, such as RK2 and RK4. However, these methods require very small post-integration slices of the time variable, which is equivalent to having many layers of ResNet and those can lead to high Memory cost. That's not what the introduction of differential equations was about. For example, when using the Euler method to solve Equation \ref{eq:3}, after K-step iterations we get,

\begin{equation} 
\begin{split}
\boldsymbol{\mathrm{h}}_1 &:= \boldsymbol{\mathrm{h}}_0 + f(\boldsymbol{\mathrm{h}}_0) \\
&\cdots\\
\boldsymbol{\mathrm{h}}_k &:= \boldsymbol{\mathrm{h}}_{k-1} + f(\boldsymbol{\mathrm{h}}_{k-1}) \\
\end{split}
\end{equation}
Which is similar to having $k$ blocks of ResNet.  If $k$ is 1M, it would be ResNet with 1M layers and will cause memory issues. Neural ODE introduced Adjoint method to solved the issues. The Adjoint method is the introduced second time backward ODE that keeps track on the gradient at time $t$ and then backpropagates with the gradient at time $t$. Since the gradient at any time can be obtained from the integral, the memory issues can be solved. For example, we have the following loss function evaluating from time $t_0$ to $t_1$ with parameters $\theta_t$,
\begin{equation} 
L (h(t_1)) = L (\int_{t_0}^{t_1}f(h(t),t,\theta)dt) = L (ODESolve(h(t_0),f,t_0,t_1,\theta))
\end{equation}
We can compute the gradient of $L$ w.r.t. hidden state with infinitesimal change and define it as Adjoint state,
\begin{equation} 
a(t) = - \frac{\partial L}{\partial \mathrm{h}(t)}
\end{equation}
It's derivative on time t, which describes the dynamics of Adjoint state is given by,
\begin{equation} 
\frac{da(t)}{dt} = -a(t)^T \frac{\partial f(t,\mathrm{h}(t),\theta_t)}{\partial \mathrm{h}(t)}
\end{equation}
It is also an ODE and its solution can also be written in integral form as follows,
\begin{equation} 
a(t) = \int a(t)^T \frac{\partial f(t,\mathrm{h}(t),\theta_t)}{\partial \mathrm{h}(t)} dt
\end{equation}
Numerical solutions at different time $t$ can be obtained by an ODE solver. The gradient at any time t can be obtained by invoking the ODE solver backwards in time from the initial point which is the gradient at time $t_1$(the gradient of the loss function on the output layer and it is easy to compute), e.g. the gradient at time $t_0$ can be solved as follows,
\begin{equation} 
a(t_0) = \int_{t_1}^{t_0} -a(t)^T \frac{\partial f(t,\mathrm{h}(t),\theta_t)}{\partial \mathrm{h}(t)} dt
\end{equation}
Similarly, we can compute the gradient of loss function w.r.t. parameters $\theta$,
\begin{equation} 
\frac{dL}{d\theta} = \int_{t_1}^{t_0} -a(t)^T \frac{\partial f(t,\mathrm{h}(t),\theta_t)}{\partial \theta} dt
\end{equation}
It can also be solved by an ODE solver and all three integrals can be solved with an ODE solver by vectorising the problem. 
\subsection{Continuous Normalizing Flow and SEMs}
We assume the assignments of SEMs are ANMs. Therefore, we can train a model which transform $N_j$ from simple distribution to input data $X$,
\begin{equation} \label{eq:351} 
\begin{split}
Z_j^{(0)} &:= N_j \\
Z_j^{(t)} &:= X_j^{(t)} - f_j(X_{\pi_j^G}^{(t)})\\
Z_j^{(1)} &:= X_j \\
\end{split}
\end{equation}
where $t$ is state variable in model, which can be $t$ hidden layer in neural networks or $t$ block in normalizing flows. We also can have residual form of equation (\ref{eq:351}) as follows,
\begin{equation}
\begin{split}
Z_j^{(0)} &:= N_j \\
Z_j^{(t+1)} &:= Z_j^{(t)} + g_j(Z_j^{(t)}), \hspace{0.5cm} \text{where} \hspace{0.2cm} g_j(Z_j^{(t)}) = \Delta\left[X_j^{(t)} - f_j(X_{\pi_j^G}^{(t)})\right]\\
Z_j^{(1)} &:= X_j \\
\end{split}
\end{equation}
If we continuously add more blocks or layers to a limit and we can have the continuous dynamics of $Z_j^{(t)}$ with an ordinary differential equation(ODE)\cite{chen2018} parameterized by $\theta$,

\begin{equation} \label{eq:353}
\frac{dZ(t)}{dt} = f(Z(t),t,\theta) 
\end{equation}
The equation (\ref{eq:353}) can be solved by a black box of ODE solver and this continuous dynamics models called Continuous Normalizing Flows(CNF\cite{ode2018}). The change of log density is also a differential equation name Instantaneous Change of Variables\cite{ode2018},
\begin{equation} 
\frac{\partial \log p(Z(t))}{\partial t} = -\mathrm{Tr}\hspace{1pt}\left(\frac{df}{dZ(t)}\right) 
\end{equation}
Therefore, the change from $Z(0)$ to $Z(1)$ can be computed by,

\begin{equation} 
\log p(Z(t_1)) = \log p(Z(t_0)) - \int_{t_0}^{t_1}\mathrm{Tr}\hspace{1pt}\left(\frac{df}{dZ(t)}\right) dt 
\end{equation}
which is the log function we try to maximize. We can solve the integral with a ODE solver and backpropagate the solution with the Adjoint Method(\cite{ffjord2019})
\section{Acyclicity Constraint and Jacobian Matrix}
\subsection{Linear Case: NOTEAR's Acyclicity Constraint}
We consider a linear case of SEM in NOTEAR\cite{zheng2018}, which has the form of $f_j(X) = W_j^T(X)$. We define $W=[W_1|W_2|\cdots|W_d] \in \mathbb{R}^{d\times d}$ is the coefficient matrix which encodes a graph. When $W_{ij} = 0$ then there is no edges from node $i$ to node $j$, when $W_{ij} \ne 0$ there exists a edge from node $i$ to node $j$ in the graph. NOTEAR proposed that if the graph is directed acyclic, then the following condition should to be satisfied,
\begin{equation}  
h(W) = \mathrm{Tr}(e^{W\circ W}) - d = 0
\end{equation}
where $\circ$ is for Hadamard product, $\mathrm{Tr}$ is for trace function of matrix and $e^M=\sum_{k=0}^{\inf}\frac{M^k}{k!}$. Let us see why this constraint can express the condition of a directed acyclicity. If the element $(i,j)$ in the $k$-th power of a non-negative adjacency matrix A $(A^k)_{ij} > 0$ , then there exists a path of length $k$ between node $i$ and node $j$. If the element (i,i) in the $k$-th power is greater than 0, then there exists a cycle in the graph. The zero power has a value of 1, then the exponential power of matrix A must be $d$ which is the dimension of data to ensure that the graph is a DAG. Also to ensure non-negativity, Hadamard product can be used. And it is easy to calculate the gradient of $h(W)$ by the following equation,
\begin{equation} 
\nabla  h(W) = (e^{W\circ W})^T \circ 2W 
\end{equation}
Meanwhile, we can use the equation as follows to simplify the calculation,
\begin{equation} 
h(W) = \mathrm{Tr}[(I+\alpha W\circ W)^d] - d = 0
\end{equation}
Where $\alpha$ can be any value greater than 0 and gradient computation can be done by deep learning framworks such as Pytorch's Autograd rather than being written manually in code implementation. 

\subsection{Non-Linear Case: Jacobian Matrix and Acyclicity Constraint}
However, In nonlinear SEM cases, we cannot find a linear $W$ and we can use partial derivatives to represent the causal dependency of $f_j$ on the $k$th variable. We denote the partial derivatives of of $f_j$ on the $k$th variable by $\partial_kf_j$ and there exits a edge from node $j$ to $k$ if and only if $\partial_kf_j\ne0$. Therefore, the Jacobian matrix $J$ represents causal dependencies between input variables $X_1,X_2,\cdots,X_n$ and $h(W)$ in nonlinear SEM cases is,
\begin{equation}  
h(J) = \mathrm{Tr}(e^{J\circ J}) - d = 0
\end{equation}
It's also easy to get that $J$ equals $W$ in linear cases, so it can also be argued that $W$ is only a special case of $J$.
\subsection{Augmented Lagrangian Optimization}
And now, the maximum likelihood optimization problems we need to solve is,
\begin{equation} 
\log p(Z(t_1)) = \log p(Z(t_0)) - \int_{t_0}^{t_1}\mathrm{Tr}\hspace{1pt}\left(\frac{df}{dZ(t)}\right) dt \hspace{0.2cm} \textrm{s.t.} \hspace{0.2cm} h(J)=0
\end{equation}
We can use the Augmented Lagrangian method to solve this optimization problem. The Augmented Lagrangian method adds a quadratic penalty to the Lagrangian method so that the converted problem can be solved more easily. Therefore, the maximum likelihood optimization problem can be transformed with Augmented Lagrangian method as\cite{yu2019},
\begin{equation} 
L(J,\theta,\lambda) = \log p(Z(t_1)\mid\theta,J) - \frac{\rho}{2}|h(J)|^2 - \lambda h(J)
\end{equation}
where $\rho$ and $\lambda$ are quadratic penalty coefficient and Lagrangian multiplier respectively.  When $\rho$ is sufficiently large,$J_*$ and $\theta_*$ are minimum point of the loss function, and the parameters obtained must satisfy $h(J)=0$. Therefore, we incrementally increase the value of $\rho$ and then optimize the entire neural network under this condition, while updating the Lagrange multiplier $\lambda$ accordingly to make it converge to the optimal point. 
\section{Related Work}
Traditionally, there are three main families of methods for causal structure learning, namely, constraint-based methods, score-based methods and structural causal function model-based methods. Constraint-based methods use conditional independence test between variables to determine a particular structure and then determine the direction based on a particular V-structure\cite{Peter2017}. The score based approach uses a score function to search for the optimal network structure and is the basis of the methodology of this paper. The structural causal model-based approach is based on structural causal model of the data generating mechanism and extends the structural causal model to increase the expressive power to discover the causal relationship between variables. 
\subsection{Constraint-based Methods}
Constraint-based methods are used to learn a set of causal networks that satisfy the conditional independence between variables in data. We use statistical test methods to verify that candidate causal networks satisfy the Causal Faithfulness Assumption.

\textbf{Definition of Causal Faithfulness Assumption\cite{Peter2017}} Given the set of variables $Z$, variables $X_i$ and $X_j$ are independent of each other or conditionally independent, then all paths between variables $X_i$ and $X_j$ are $d$-separated by the set of variables $Z$ in the causal graph $\mathbb{G}$ that defines the process by which data $X$ is generated. Then the joint distribution $P_X$ over random variables $X$ is Causal Faithfulness to the graph $\mathbb{G}$.

There are three steps in this family of algorithms, the skeleton learning stage, direction learning stag and possible orientation stage. In the learning phase of the skeleton graph, an skeleton graph without orientations is obtained by the independence of the variables with independence tests or conditional independence tests technologies. Commonly used tests for conditional independence are the statistical analysis-based chi-square test or the information theory-based mutual information test. In the direction learning phase, direction is determined based on a specific V-structure. In the possible orientation stage, we use three rules to orient undirected edges as many as possible. The main problem with this family of methods is that the number of conditional independent test grows exponentially as the number of nodes increases, and the computational cost is very high. So the main research direction of such algorithms is to reduce the number of tests.

We briefly introduce the Peter Clark(PC\cite{PC2000}) algorithm here. At the first stage, the skeleton of DAG with undirected edges estimated. We start with a completed connected graph with no oriented edges and search depth equals 0(depth=0 means the neighbour nodes of test nodes). For each pair of nodes $X_i$ and $X_j$, test one by one that given neighbor node $X_k$ of the two in the graph, whether these two nodes are conditionally independent. If yes, then remove the edge of these two nodes $X_i$ and $X_j$ and add neighbor node $X_k$ to the set of $d$-separated $S_{ij}$. When all edges are removed with depth=0, increase the depth to 1 and repeat this process until the number of neighbors of the node is less than the depth. In the second stage of PC algorithm, For each pair of unconnected nodes $X_i$ and $X_j$ with a common connected neighbour $X_k$, if $X_k$ is not in $d$-separate set $S_{ij}$ then the undirected V-Structure $X_i - X_k - X_j$ is orientated to $X_i \rightarrow X_k \leftarrow X_j$. Otherwise $X_k$ is not a collider of the V-Structure. In the third stage, we continue to check if there is new edges can be oriented with three rules avoiding new V-Structure 
discovered and new cycles(the graph is acyclic): (1) we point from $X_i$ to $X_j$ if $X_k$ pointing to $X_j$ and $X_i$ is not the neighbour node of $X_i$; (2) we point from $X_i$ to $X_j$ if there exists a chain $X_i \rightarrow X_k \rightarrow X_j$; (3) we point from $X_i$ to $X_j$ if $X_i - X_k \rightarrow X_j$ and $X_i - X_l \rightarrow X_j$.

The Inductive Causation(IC\cite{IC1995}) algorithm and its variants\cite{ICE1995} are similar to the PC algorithm in that they also use three stages to learn the causal network structure. However, most independence tests are chi-square test or partial correlation tests based on Gaussian distribution or multinomial distribution. To overcome these limitations, many effective methods have been proposed to handle more complex data distributions. For example, using Kernel based Hilbert-Schmidt Norms and Kernel-base conditional independence test for more complex distributed data. Further more, when Causal Faithfulness Assumption is violated, there may be unobservable confounding factors. The FCI(Fast Causal Inference\cite{FCI1993}) algorithm and FCI improved RFCI (Really Fast Causal Inference\cite{RFCI2012}) algorithm are proposed to the discovery of causality with hidden variables through extended graphs. 

Constraint-based methods are effective for discovering causality and can be widely used with given reliable conditional independence tests. However, it is not possible to determine the direction of all edges through conditional independence tests and V-structures. Therefore, we need other types of methods to do causal learning.

\subsection{Scored-based Methods}
Score-based methods are an alternative to learning causal structures. A score-based approach uses a scoring function to quantify how well a Bayesian network fits a given distribution of data and then uses a search algorithm to find the graph structure that best fits the data. In this approach, the choice of the scoring function is crucial, the scoring function maps the candidate causal graph to a certain scalar based on a given structure. Bayesian Information Criterion(BIC\cite{BIC1978}) is commonly and widely used one and its formula is $BIC(X,G)=k\ln(n)-2\ln(L)$, where $L$ is the maximized value of the likelihood function of given graph $G$ and n is number of the samples and $k$ denotes the number of the variables. However, BIC failed to do feature selection in high-dimension data. Another popular one of the  Bayesian score function is the Bayesian Dirichlet equivalent uniform
(BDeu\cite{BeU1995}) score which has the form of,
$$
S(X,G)=\prod_{i=1}^{n}\prod_{j=1}^{q_i}\frac{\Gamma(\alpha/q_i)}{\Gamma(\alpha/q_i + \sum_{k=1}^{r_i}s_{ijk})}\prod_{k=1}^{r_i}\frac{\Gamma(\alpha/q_i + s_{ijk})}{\Gamma(\alpha/q_i)}
$$
where $r_i$ is the number of stats of $X_i$; $q_i$ indicate the number of configurations of the parents of $X_i$; $s_{ijk}$ denotes the number of observation data that $X_i$ is for its $k$-th value and the parents of $X_i$ took $j$-th sample. 

However, the number of candidature graph structures grows exponentially as the number of variables increases and the problem becomes NP-hard due to the large search space. Therefore, heuristic search algorithms such as Greedy Equivalence Search(GES\cite{Meek1997}) and its extension Fast GES(FGES\cite{Ramsey2016}) are often used to find a locally optimal graph. In GES algorithm, there are two stages, a forward phase where edges are added and a backward phase where edges are removed. In the forward phase, edges are added in a greedy manner (i.e., maximizing the score which is calculated by a score function defined by GES) until  
score can not be further increased. In the second phase, the edges are greedily removed until the score is optimal. GES can search the graph space in a very efficient way because it includes a greedy algorithm. However, the scoring process of the algorithm is too redundant and adding edges causes the number of scoring to increase exponentially. It means that adding edges can make the time complexity grow exponentially and it becomes impractical as the number of variables increases. The FEGS algorithm improves the GES algorithm by decreasing the computational complexity when adding a new edge. Moreover, FEGS parallelize special steps and does not depend on the order of operations, which makes the scoring processes much faster than GES algorithm.

The hybrid approach combines Scored-based Methods and Constraint-based methods to overcome their respective drawbacks by using conditional independence tests to reduce the complexity of the candidate graph search space, followed by a scoring-based approach to find the best network structure. For example, the Max-Min Hill-Climbing(MMHC\cite{Tsamardinos2006}) algorithm first learns a skeleton of graph by the Max-Min Parents and Children(MMPC\cite{Tsamardinos2003}) algorithm, which is equivalent to a constraint-based approach, followed by a greedy Bayesian score climbing search method to orient the graphs. This approach is not only suitable for high-dimensional data, but also improves the effectiveness of learning causal structures.
\subsection{Structural Causal Function Model-based Methods}
Constraint-based methods have Markov equivalence class problems and cannot orient all edges while score-based methods are not efficient due to the large search space, therefore, many studies have proposed structural causal models from the perspective of data generation or causal mechanisms between the variables of data. The general form of the structural causal model has the form of $X_j := f_j(X_{pa_j}, N_j)$ Where $X_{pa_j}$ is for the set of parent nodes of $X_j$ and $N_j$ is for mutually independent noise. The structural causal model describes the mechanism for generating data between variables rather than an algebraic equation describing the equality of left and right sides. However, different SEMs to entail a same distribution $P_X$ on $X(X_1,X_2,\cdots,X_d)$. Therefore, more information such as stronger assumption of the data generation method $f_j$ should be provided. These algorithms with stronger assumptions include Linear Non-Gaussian Acyclic Model(LiNGAM), Post-NonLinear(PNL), Additive Noise Model(ANM) in non-linear cases and its extensions, Information-Geometric Causal Inference(IGCI) and hybrid algorithms combining Constraint-based methods and Structural Causal Function Model-based Methods. The detail of LiNGAM, ANM, LGMEER can be seen at subsection of \ref{scms}    
\section{Experiments}
In this section we experimentally verify whether it is possible to derive dynamic physical systems from observational data via Neural ODEs and then read the causal structure between variables in the data from the physical systems. Deriving dynamic physical systems with SCMs from observational data has not been done before, so it is not possible to compare with previous works. But previous works of learning the causal structure between variables allows us to get some benchmarks. The aim of this paper is to verify the idea of obtaining a dynamic physical system from observational data and then reading out the causal structure of the variables in the data from the physical system. However, for learning the causal structure of the variables, this approach outperforms previous works to learn the causal structure in some datasets.

\textbf{Baselines} we choose the following algorithms as baselines for comparison: two gradient-based methods GraN-DAG\cite{{gran2020}} and Sparse-DAG\cite{sparse2020} using weights in neural networks as causal dependencies; CAM\cite{cam2014} for non-linear additive structural causal models based method; NOTEARS for linear structural causal models and its non-linear extension DAG-GNN\cite{yu2019}. Other algorithms such as PC, GES and FGS have been shown to be poor performance in multiple experiments\cite{gran2020,sparse2020,yu2019}, so we omitted.

\textbf{Metrics} we choose the following metrics to evaluate the causal structure learned observational data: True Positive Rate(TPR) and the structural hamming distance. The former is the number of correctly identified oriented edges divided by the total number of oriented edges in true DAG and the latter counts the number of falsely adding, deleting and orienting edges. 

\subsection{Synthetic Data}
In the synthetic data experiments, we used Erdös–Rényi(ER) as the graph type to generate random graphs $G$ and generated data from the random graphs $G$ in which the causal order defined. We generated datasets $X_1,X_2,\cdots,X_d$ with $d=10$ and $1d$ and $4d$ edges denoted by $ER1$ and $ER4$ respectively. The data generating process we choose is Non-linear Gaussian ANM with the form of $  X_j := f_j(X_{pa_j}) +  N_j \hspace{0.2cm} j=1,...d, $ 
Where $X_{pa_j}$ is for the set of parent nodes of $X_j$ and $N_j$ is for mutually independent unit Gaussian noise and $f_i$ we used is Gaussian Process(GP) with a unit bandwidth RBF kernel. Due to non-linear assignment of $f_j$ and Gaussian noise, the DAG is identifiable from the distribution $P_X$ over data $X$. The results of comparisons among different methods are showed in Table \ref{tab:2}, in which we can that our proposal method DAG-ODE outperforms others algorithms in any aspects. 

\subsection{Real Data}
We evaluate the real dataset that is generally accepted by the biological community and is often used as a benchmark. The data consists of 11 continuous variables corresponding to different proteins and phospholipids in cells of the human immune system and 7466 observations, each of which indicates the measured level of each biological molecule in a single cell under different experimental interventions\cite{{sachs2005}}. 
\begin{figure}
    \centering
    \includegraphics[width=14cm]{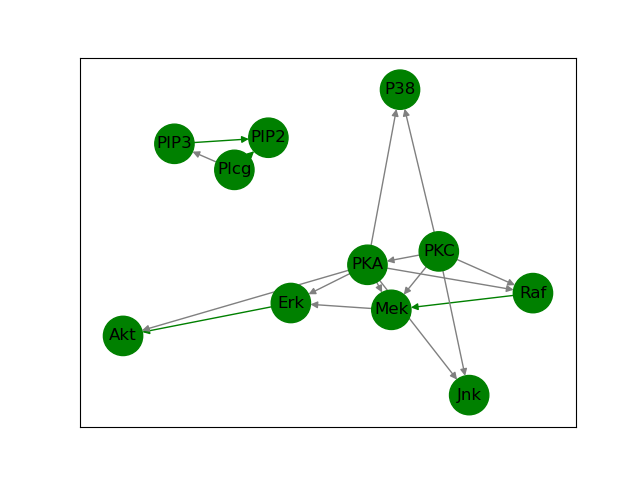}
    \caption{The causal graph of Sachs dataset estimated by our methods, in which the gray arrows represent missing edges from the groundtruth.}
    \label{fig:sachs3}
\end{figure}
While the groundtruth of the The consensus network is 17 edges, we report SHD of 13 estimated 4 edges which are all expected edges as shown in Figure \ref{fig:sachs3}. For detail, the 4 true positives are Raf $\rightarrow$ Mek, Plcg $\rightarrow$ PIP2, PIP3 $\rightarrow$ PIP2, Erk $\rightarrow$ Akt. By comparison, while DAG-GNN reports SHD of 19 with 18 edges predicted, GraN-DAG estimated 16 edges with SHD of 13 and Sparce-DAG predicted 13 edges with SHD of 16.

\section{Conclusion}
In this work, we extend jacobian-based to physical system which is the method human explore and reason the world and it is the highest level of causality. By functions fitting with Neural ODE, we can read out causal structure from functions. Our approach also enforces a important acylicity constraint on continuous adjacency matrix of graph nodes and significantly reduce the computational complexity of search space of graph. For the task of structure learning, our method outperforms other current state-of-art methods for learning causal structures in experiments of datasets of 10 nodes and improves the performance in datasets with more dense causal relationships.

\bibliographystyle{unsrt}  


\end{document}